\documentclass[letterpaper]{article} %
\usepackage{aaai25} %
\usepackage{times}  %
\usepackage{helvet}  %
\usepackage{courier}  %
\usepackage[hyphens]{url}  %
\usepackage{graphicx} %
\usepackage{natbib}  %
\usepackage{caption} %
\usepackage{algorithm}
\usepackage{algorithmic}
\usepackage{booktabs}
\usepackage{newfloat}
\usepackage{listings}
\DeclareCaptionStyle{ruled}{labelfont=normalfont,labelsep=colon,strut=off} %
\floatstyle{ruled}
\newfloat{listing}{tb}{lst}{}
\floatname{listing}{Listing}
\usepackage{url}
\usepackage[dvipsnames]{xcolor}
\usepackage{tabularx}
\usepackage{graphicx}
\usepackage{mathtools}
\usepackage{amsthm}
\usepackage{amssymb}
\usepackage{amsmath}
\usepackage{multirow}
\usepackage{breqn}
\usepackage{todonotes}
\usepackage{soul}
\usepackage{caption}
\usepackage{subcaption}
\usetikzlibrary{calc,matrix}

\DeclarePairedDelimiter\multiset{\lbrace\!\!\lbrace}{\rbrace\!\!\rbrace}
\DeclareMathOperator*{\clip}{clip}

\DeclareMathOperator{\Agg}{\text{\textsc{Aggregate}}}
\DeclareMathOperator{\Comb}{\text{\textsc{Combine}}}

\DeclareMathOperator{\Readout}{\text{\textsc{ReadOut}}}

\DeclareMathOperator{\MLP}{\mathrm{MLP}}

\DeclareMathOperator*{\argmax}{arg\,max}
\newcommand{\argtopk}{\mathop{\mathrm{arg\,top}\,k}}

\newcommand{\LK}[1]{\textcolor{black}{#1}}

\nocopyright
\title{Crossfire: An Elastic Defense Framework for \\ Graph Neural Networks under Bit Flip Attacks}

\author{
    Lorenz Kummer\textsuperscript{\rm 1,\rm 2}, Samir Moustafa\textsuperscript{\rm 1,\rm 2}, Wilfried N. Gansterer\textsuperscript{\rm 1}, Nils Kriege\textsuperscript{\rm 1,\rm 3}
}

\affiliations {
    \textsuperscript{\rm 1}Faculty of Computer Science, University of Vienna, Vienna, Austria\\
    \textsuperscript{\rm 2}Doctoral School Computer Science, University of Vienna, Vienna, Austria\\
    \textsuperscript{\rm 3}Research Network Data Science  University of Vienna, Vienna, Austria\\
    \{lorenz.kummer, samir.moustafa, wilfried.gansterer, nils.kriege\}@univie.ac.at
}

\usepackage{bibentry}

\begin{document}

\maketitle

\begin{abstract}
Bit Flip Attacks (BFAs) are a well-established class of adversarial attacks, originally developed for Convolutional Neural Networks within the computer vision domain. Most recently, these attacks have been extended to target Graph Neural Networks (GNNs), revealing significant vulnerabilities. This new development naturally raises questions about the best strategies to defend GNNs against BFAs, a challenge for which no solutions currently exist. Given the applications of GNNs in critical fields, any defense mechanism must not only maintain network performance, but also verifiably restore the network to its pre-attack state. Verifiably restoring the network to its pre-attack state also eliminates the need for costly evaluations on test data to ensure network quality. %
We offer first insights into the effectiveness of existing honeypot- and hashing-based defenses against BFAs adapted from the computer vision domain to GNNs, and characterize the shortcomings of these approaches. %
To overcome their limitations, we propose Crossfire, a hybrid approach that %
exploits weight sparsity and combines hashing and honeypots with bit-level correction of out-of-distribution weight elements to restore network integrity. Crossfire is retraining-free and does not require labeled data. Averaged over 2,160 experiments on six benchmark datasets, Crossfire offers a 21.8\% higher probability than its competitors of reconstructing a GNN attacked by a BFA to its pre-attack state. These experiments cover up to 55 bit flips from various attacks. Moreover, it improves post-repair %
prediction quality by 10.85\%. Computational and storage overheads are negligible compared to the inherent complexity of even the simplest GNNs.
\end{abstract}

\section{Introduction}
Graph Neural Networks (GNNs) are effective machine learning methods for processing structured data in graph format, consisting of nodes and edges. They demonstrate versatility by enabling the application of deep learning in diverse domains such as finance, social networks, medicine, chemistry, and biological data analysis~\cite{lu2021weighted, cheung2020graph, sun2021disease, 
wu2018moleculenet, xiong2021novo}.
As GNNs become more widely used, it is crucial to examine their potential security vulnerabilities. Conventional adversarial attacks on GNNs primarily involve manipulating input graph data~\cite{wu2022graph}. These attacks include poisoning, which leads to the learning of flawed models~\cite{ma2020towards, wu2022graph}, and evasion strategies, which use adversarial examples to impair inference.
Such attacks on GNNs, which involve altering node features, edges, or introducing new nodes~\cite{sun2019node, wu2022graph}, as discussed in prior studies~\cite{%
ma2020towards}, can be either targeted or untargeted. Targeted attacks reduce the model's prediction quality on specific instances, whereas untargeted attacks affect the model's overall performance~\cite{zhang2022unsupervised}. For a thorough overview of %
graph poisoning and evasion attacks, along with defenses and relevant algorithms, refer to the detailed reviews by \citet{jin2021adversarial} and \citet{dai2022comprehensive}.

Recent research on GNN vulnerability has expanded beyond poisoning and evasion attacks on input graphs, now encompassing systematic attacks that directly manipulate network weights through malicious bit flips during inference. While such attacks
are well-understood for Convolutional Neural Networks (CNNs)
in computer vision~\cite{rakin2019bitflip,
qian2023survey, khare20222design}, understanding GNN vulnerability to BFAs is a relatively new area of study. %
Currently, only one dedicated BFA for GNNs has been described by~\citet{kummer2023ibfa}, and existing defenses against BFAs~\cite{khare20222design} have not been evaluated for GNNs. %
To the best of our knowledge, the topic of defending GNNs against BFAs has not yet been addressed in the literature.

Given the fundamental differences between GNNs and CNNs -- such as GNNs' reliance on the message-passing algorithm (MP) to process graph-structured data and the lack of vulnerable convolutional filters~\cite{hector2022evaluating} -- as well as distinct properties like expressivity that can be exploited by attackers~\cite{kummer2023ibfa}, it is essential to explore whether existing CNN defenses against BFAs are applicable to GNNs.
Moreover, it is crucial to develop defenses specifically designed for GNNs, given the importance of their applications across multiple domains. 

In the following, we assume a white-box threat model to evaluate CNN defenses against BFAs on GNNs and introduce our approach, considering an attacker who can precisely manipulate bits without budget constraints.

\paragraph{Related Work}
The BFA initially presented in \citet{rakin2019bitflip}, targets a quantized CNN by conducting a single forward-backward pass using a randomly selected training data batch, without updating the weights. It identifies the top-$k$ binary gradient bits as potential bit-flip candidates. These bits are then flipped iteratively to maximize the loss (using the same loss function as in training) until the desired network degradation is achieved. This original BFA is termed Progressive BFA (PBFA). We use BFA to refer to a broader range of attacks that systematically induce malicious bit flips in a neural network's weights and biases, and refer the interested reader to the comprehensive survey for BFAs on CNNs by \citet{qian2023survey}. For GNNs, only a single dedicated BFA has been explored so far, termed Injectivity Bit Flip Attack (IBFA), which exploits certain properties related to GNN expressivity by~\citet{kummer2023ibfa}.

To defend against BFAs on CNNs, several approaches have been proposed. Network hardening methods, such as adversarial~\cite{he2020defending, he2020defending2} and perturbation-resilient training~\cite{chitsaz2023training}, as well as gradient obfuscation strategies \cite{zhang2022mitigating, wang2023aegis}
, focus on resisting gradient-based bit search attacks. However,  they may require (re)training, which limits their ex-post deployment, and they do not necessarily rectify compromised neurons. Detection-focused methods leverage hashing~\cite{java2021hashtag, li2021radar} %
or output code matching~\cite{ozdenizci2022improving} for efficient bit flip detection and consistently achieve high detection rates in the computer vision domain. These methods often wrap around existing pre-trained networks and can be applied ex-post. However, their restoration capabilities are limited, often necessitating retraining or pruning of manipulated weights. Proactive defenses take a different approach by anticipating attacks through analyzing assumed underlying attack mechanisms \cite{liu2023neuropots, he2020defending2}. These methods prioritize weight restoration over pruning, aiming to revert the network to its pre-attack configuration \cite{liu2023neuropots}, or use statistical approximation to address compromised weights~\cite{he2020defending2}. For an %
overview of defense mechanisms, see the review by \citet{khare20222design}.

\paragraph{Challenges and Limitations}
Existing defenses against BFAs face shared challenges and limitations. First, they lack mechanisms to ensure full network restoration after attacks; while prediction quality may recover, network integrity is not guaranteed. Reloading a clean model or delegating control to a backup system after attack detection is often inefficient and delay-prone~\cite{li2021radar}. Second, reliance on a single defense mechanism leaves these approaches susceptible to loophole-exploiting BFAs~\cite{liu2023neuropots}.

Moreover, the transferability of defense strategies from CNNs to GNNs remains unexplored. Given GNNs' critical roles in applications like medical diagnosis~\cite{li2020graph, %
lu2021weighted}, health record modeling~\cite{liu2020heterogeneous, sun2021disease}, and drug development~\cite{
xiong2021novo, cheung2020graph}, ensuring their authenticity post-attack is essential. A robust defense strategy must integrate multiple protective layers and incorporate efficient, low-cost verification.

\section{Contribution}
\textbf{(1)} We study %
two state-of-the-art BFA defense methods based on honeypots \cite{liu2023neuropots} and hashing \cite{li2021radar}, which were initially developed for CNNs, and assess their effectiveness for GNNs, identifying notable weaknesses in their application to this domain. \textbf{(2)} We propose an innovative honeypot selection method using unlabeled data to enhance detection rates and validate GNN reconstruction using a robust, well-known and studied, yet lightweight hash function. \textbf{(3)} Moreover, we build upon existing work and introduce an efficient weight group-based checksum approach to detect hits in non-honeypot weights, addressing the limitation of solely relying on honeypots. This approach, together with the bit-level correction of out-of-distribution (OOD) weight elements as well as the careful exploitation of the relationship between BFAs and GNN weight sparsity, allows us to reconstruct the network in certain cases even when hits on the preselected honeypots are avoided by a diligent attacker.

Our work establishes a strong and dependable defense framework, obviating the need for post-reconstruction quality verification of the GNN.

\section{Preliminaries}
This section provides an overview and detailed description of the key elements essential to our research: GNNs, BFAs, the threat scenario, as well as the honeypot-based NeuroPots~\cite{liu2023neuropots} and the hashing-based runtime Adversarial Weight Attack Detection and Accuracy Recovery (RADAR) defense methods~\cite{li2021radar}.

\paragraph{Graph Neural Networks}
GNNs leverage graph structure and node features to derive representation vectors for specific nodes, denoted as $\mathbf{h}_v$ for node $v$, or for the entire graph $G$, denoted as $\mathbf{h}_G$. Contemporary GNNs use a neighborhood aggregation or message-passing approach, where the representation of a node is iteratively updated by aggregating the representations of its neighboring nodes.
Upon completion of $k$ layers of aggregation, the representation of a node encapsulates the structural information within its $k$-hop neighborhood~\cite{leskovec2019powerful, welling2016semi}. The $k$th layer of a GNN computes the node features $\mathbf{h}_v^{k}$ %
as defined by
\begin{align}
     \nonumber
     \mathbf{a}_v^{k} &= \Agg^{k}\left ( \multiset{ \mathbf{h}_u^{k-1} \mid u \in N(v)} \right )
    \label{eq:defgnn1} 
    \\
     \mathbf{h}_v^{k} &= \Comb^{k}\left ( \mathbf{h}_v^{k-1}, \mathbf{a}_v^{k} \right )
    \nonumber
\end{align}
with node neighborhood $N(v)$ and multiset $\multiset{ }$.
Initially, $\mathbf{h}_v^{0}$ represents the features of node $v$ in the given graph. As the choice of $\Agg^{k}$ and $\Comb^{k}$ in GNNs is critical, several variants have been proposed~\cite{leskovec2019powerful}.

\paragraph{Quantization}
Quantization reduces the precision or adopts efficient representations for weights, biases, and activations, decreasing model size and memory usage~\cite{benoi2018quantization, kummer2023adapt}.

In accordance with the typical setup chosen in related work on BFA, we apply scale quantization to map \texttt{FLOAT32} tensors to the \texttt{INT8} range.
Such a quantization function $\mathcal{Q}$ and its associated dequantization function $\mathcal{Q}^{-1}$ are
\begin{equation}
    \begin{aligned}
        \mathcal{Q}(\mathbf{W}^{l}) &=  \mathbf{W}_{q}^{l} = \clip(\lfloor \mathbf{W}^{l} / s \rceil, \; a, b), 
        \\ 
        \mathcal{Q}^{-1}(\mathbf{W}_{q}^{l}) &= \widehat{\mathbf{W}}^{l} = \mathbf{W}_{q}^{l} \times s. 
    \end{aligned}
    \label{equation:neural_network_qauntizer}
\end{equation}
Here, $s$ denotes the scaling parameter, $\clip(x, a, b) = \min(\max(x,a),b)$ with $a$ and $b$ the minimum and maximum thresholds (also known as the quantization range), $\lfloor \dots \rceil$ denotes nearest integer rounding, $\mathbf{W}^{l}$ is the weight of a layer $l$ to be quantized, $\mathbf{W}_{q}^{l}$ its quantized counterpart and $\widehat{\dots}$ indicates a perturbation, i.e., rounding errors in the case of~\eqref{equation:neural_network_qauntizer}.
Similar to other works on BFA that require quantized target networks such as~\cite{rakin2019bitflip}, we address this issue of non-differentiable rounding and clipping functions present in $\mathcal{Q}$
by using Straight Through Estimation (STE)~\cite{Bengio2013EstimatingOP}.
\paragraph{Bit Flip Attacks}
PBFA, introduced in the seminal work by \citet{rakin2019bitflip}, uses a quantized trained CNN $\Phi$ and progressive bit search (PBS) to identify bits for flipping. PBS starts with a forward and backward pass, performing error backpropagation without updating weights on a randomly selected batch $\mathbf{X}$ of training data with a target vector $\mathbf{t}$. It then selects the weights corresponding to the top-$k$ largest binary encoded gradients as potential candidates for bit flipping. These candidate bits are iteratively tested across all $L$ layers to find the bit that maximizes the difference between the loss $\mathcal{L}$ of the perturbed CNN and the loss of the unperturbed CNN,
\begin{equation}
    \begin{aligned}
            \max_{\{ \widehat{\mathbf{W}}_{q}^{l} \}} & \mathcal{L}\Big (\Phi \big( \mathbf{X}; \; \{\widehat{\mathbf{W}}_{q}^{l}\}_{l=1}^{L} \big), \; \mathbf{t} \Big) - \mathcal{L}\Big (\Phi \big( \mathbf{X}; \; \{\mathbf{W}_{q}^{l}\}_{l=1}^{L} \big), \; \mathbf{t} \Big) %
    \end{aligned}
    \label{eq:bfa_optimization}
    \nonumber
\end{equation}
whereby the same %
$\mathcal{L}$ is used that was minimized during training, e.g., (binary) cross entropy (B)CE for (binary) classification). 
Using \( \ell_1 \) or Kullback-Leibler-Divergence~\citep{kullback1951information} for $\mathcal{L}$, the variant IBFA by~\citet{kummer2023ibfa} dedicated to %
GNNs instead optimizes
\begin{equation}
    \min_{\{ \widehat{\mathbf{W}}_{q}^{l} \}} \mathcal{L}\Big (\Phi \big( \mathbf{X}_a; \{\widehat{\mathbf{W}}_{q}^{l}\}_{l=1}^L \big),  \; \Phi\big ( \mathbf{X}_b;\{\widehat{\mathbf{W}}_{q}^{l}\}_{l=1}^L ) \Big)
    \label{eq:optimibfa}
    \nonumber
\end{equation}
via PBS and uses a unique input data selection process related to its theoretical fundament given by
\begin{equation}\label{eq:find_sample}
    \argmax_{\{ \mathbf{X}_a, \mathbf{X}_b \}} \mathcal{L}\Big (\Phi \big( \mathbf{X}_a; \{\mathbf{W}_{q}^{l}\}_{l=1}^L \big),  \; \Phi\big ( \mathbf{X}_b;\{\mathbf{W}_{q}^{l}\}_{l=1}^L ) \Big).
    \nonumber
\end{equation}
IBFA targets specific mathematical properties of GNNs %
that are crucial for graph learning tasks requiring high structural expressivity. \citet{kummer2023ibfa} demonstrate IBFA's efficacy in exploiting GNN expressivity, rendering GNNs indifferent to graph structures and compromising their predictive quality in tasks that require structural discrimination.

\paragraph{NeuroPots}
\citet{liu2023neuropots} introduce NeuroPots, a proactive defense mechanism for CNNs that embeds honey neurons as %
vulnerabilities to lure attackers and facilitate fault detection and model restoration. The proportion of honey neurons per layer is controlled by the hyperparameter $p$, indicating the percentage of neurons modified. The framework uses a checksum-based detection method, anticipating that most bit flips will target these trapdoors, and uses trapdoor refreshing to recover the model's prediction quality.%

NeuroPots provides two methods: retraining-based and heuristic. The retraining approach, suitable for those with ample training data, is data-intensive. The heuristic method is more efficient, enhancing specific neuron activation for trapdoor embedding and adjusting adjacent weights to maintain its contribution to the next layer. This method is ideal for full-precision models, though it may cause minor quantization errors. It can be applied to any layer, including direct modifications of honey neuron activations in the input layer~\cite{liu2023neuropots}. The one-shot encoding process used by NeuroPots can be described as follows:
\begin{equation}
    o_i^{l+1} = \sum_{j=1}^{n^{l}} w^{l}_{ji} \cdot o^{l}_{j} = w^{l}_{0i} \cdot o^{l}_{0} + \cdots + \left( \frac{1}{\gamma} \cdot w^{l}_{hi} \right) \left( \gamma \cdot o^{l}_{h} \right)
    \nonumber
\end{equation}
where $o^{l}_h$ denotes the honey neuron at layer $l$, $w^{l}
_{hi}$ denotes the associated honey weights, and $\gamma$ the rescaling factor. For a typical neuron, its influence on the output of the next layer in the presence of BFAs can be formulated as \(o^{l+1} = (w + \Delta w) \cdot o^l\), where \(\Delta w\) denotes the weight distortion arising from bit flips. Conversely, considering the one-shot trapdoor as an example, the impact of a honey neuron on the subsequent layer can be represented as:
\begin{equation}
o^{l+1} = \left(\frac{1}{\gamma} \cdot w + \Delta w\right) \cdot \gamma \cdot o^{l} = (w + \Delta w) \cdot o^{l} + (\gamma - 1) \cdot \Delta w \cdot o^{l}
\nonumber
\end{equation}
Obviously, \(o^{l+1}\) experiences an increase of \((\gamma - 1) \cdot \Delta w \cdot o^{l}\) in comparison to a regular neuron. Furthermore, attackers tend to flip the most significant bits of weights, leading to a substantial perturbation \(\Delta w\). Consequently, the impact of the honey neuron on \(o^{l+1}\) becomes more pronounced, particularly for larger values of \(\gamma\). This alteration propagates and accumulates across subsequent layers, causing a substantial shift in the model's output. 

NeuroPots, despite its novelty, has limitations. Random honeypot selection may overlook key neurons or include inactive ones, while a global scaling parameter $\gamma$ ignores neuron-specific roles. Critically, it can not certify restoration to the pre-attack state, requiring post-recovery evaluation, as it fails to detect changes in non-honeypot neurons.

\paragraph{RADAR}
RADAR~\cite{li2021radar} is designed to protect CNN weights from PBFA. It organizes weights into groups within each layer and uses a checksum-based algorithm to generate a 2-bit to 3-bit signature per group. During runtime, this signature is computed and compared with a stored reference to detect BFAs. If detected, the weights in the affected group are zeroed %
to minimize prediction quality loss. %
RADAR is integrated into the inference computation stage.

RADAR efficiently detects most bit flips, including random ones, but only partially restores prediction quality by zeroing attacked weight groups. It cannot fully revert the network to its pre-attack state, requiring test data evaluation to confirm recovery. For large numbers of bit flips, zeroing further degrades the network and impacts non-attacked weights, causing collateral damage.

\paragraph{Threat Scenario}
In line with the general trend in literature on BFAs for CNNs, e.g.,~\citep{yao2020deephammer, rakin2019bitflip, 
he2020defending, he2020defending2, li2021radar, liu2023neuropots, java2021hashtag} as well as IBFA~\cite{kummer2023ibfa}, we assume that the target network is \texttt{INT8} quantized.
Furthermore, we adopt the assumption which is common in related work that the attacker has the capability to exactly flip the bits chosen by the bit-search algorithm through mechanisms such as RowHammer~\citep{mutlu2019rowhammer}, NetHammer~\citep{lipp2020nethammer} or others~\citep{breier2018practical%
}. For a detailed discussion on the technical aspects of implementing arbitrary flips of the identified vulnerable bits in hardware, we refer %
to~\citet{wang2023aegis}.

To test the reliability of our framework under worst-case conditions, we assume that the attacker is not subjected to budget considerations~\citep{hector2022evaluating}, although typically, flipping more than 25 bits is considered difficult for an attacker~\cite{yao2020deephammer} and 50 bits is considered an upper boundary~\cite{wang2023aegis}. Moreover, we assume that some amount of training data as well as information on the network structure is available to the attacker, which is a typical assumption in related work on BFAs, e.g.,~\citet{liu2023neuropots}. This information can be acquired through methods such as side-channel attacks~\citep{yan2020cache, batina2018csi}. \LK{Together, these typical assumptions amount to a white box threat model, whereby the attacker's goal is to crush a well-trained and deployed quantized GNN via BFA.}

In our defense strategy, we presume the presence of unlabeled data and secure storage of original honey weights and hashes in an inaccessible location for potential attackers. Similar to~\cite{liu2023neuropots}, we argue that such a secure storage can be realized via trusted execution environments (TEE) like Intel SGX~\cite{mckeen2016intel} or ARM Trustzone~\cite{pinto2019demystifying, arm2009security}.

\section{The Crossfire Defense Mechanism}
Crossfire is a rapid detection and recovery method that restores model prediction quality and verifies if the recovered model matches the pre-attack state, eliminating the need for test data. It operates in three stages: initialization, monitoring, and reconstruction. In initialization, Crossfire processes a fully trained, quantized GNN, inducing sparsity through dequantization and \( \ell_1 \) pruning. Honeypots and scaling parameters are computed from gradients and network depth, followed by re-quantization and generation of hashes for layers, rows, and columns. During monitoring, layer hashes detect alterations, triggering the reconstruction phase. In reconstruction, altered weights are identified via row and column hashes and are either restored with honeypots, corrected at the bit level (if OOD), or zeroed if the other options fail. The induced sparsity increases the likelihood that zeroing resets the GNN to its pre-attack state, with layer hashes finally verifying the GNN's integrity.

\paragraph{Inducing Sparsity}
We exploit that BFAs are prone to flip most significant bits of near-zero weights~\cite{qian2023survey, li2021radar} by inducing sparsity via \( \ell_1 \)-magnitude pruning, setting \( w \in \mathbf{W}^l \) to zero if \(|w| < \tau\), where \(\tau\) is the \( p \)-th quantile \(\Lambda_p(\left|\mathbf{W}^l\right|)\), pruning \( p \cdot 100\% \) of the smallest weights:
\[
w \gets 
\begin{cases} 
    0 & \text{if } |w| < \tau, \\
    w & \text{otherwise.} 
\end{cases}
\]
This increase in the attack surface will steer attackers toward neurons with high sparsity and help mask the most important neurons by increasing the number of potential target weights, saturating the network with zero weights. %
Theoretical insights from \cite{
frankle2018lottery, malach2020proving} suggest that introducing a limited amount of sparsity to a neural network does not significantly reduce its prediction quality. Our detailed results show that models modified by Crossfire maintain their prediction quality (see appendix).%
Further, by guiding attackers towards \( \ell_1 \)-pruned weights, subsequent checksum-based zeroing will allow reconstruction to the pre-attack state. While a sophisticated attacker might avoid targeting zeroed weights, this strategy reduces the attack's efficacy by narrowing viable options as sparsity increases, forcing less optimal bit flips.
 
\paragraph{Honeypot Selection}
Focusing on the heuristic approach due to the need for labeled data in \citet{liu2023neuropots}'s retraining method, we observe some critical aspects. Randomly selecting neuron honeypots, as proposed to counter activation-ranking attackers, ignores slight variations in neuron activation across batches. Effective deployment would thus require dynamic $\gamma$ and honeypot adjustment or a batch-averaging strategy. However, dynamic adjustment risks exposing honeypots via network changes, while batch averaging distributes honeypots across rankings, enhancing effectiveness even against advanced attackers.

Thus to improve honeypot selection, we %
employ a two-fold method:
First, we create predictions %
for a small subset of unlabeled data $S$. %
We derive classes from these %
predictions to form a set $P$ of tuples $(\mathbf{t}, \mathbf{X})$, where $\mathbf{X}$ is the input data (batch) and $\mathbf{t}$ are the predicted classes ($\mathop{\mathrm{arg\,max}}$ of class probabilities), serving as pseudo labels for backpropagation.
\begin{equation}
    \begin{aligned}
           P = \Big \{\big(\mathbf{t},\mathbf{X}\big) \mid  \mathbf{t} = \mathop{\mathrm{arg\,max}}_{axis=1}\big(\Phi \big( \mathbf{X}; \; \{\mathbf{W}^{l}\}_{l=1}^{L} \big), \mathbf{X} \in S \Big\}
    \end{aligned}
    \label{eq:make_targets}
    \nonumber
\end{equation}
For each $(\mathbf{t},\mathbf{X}\big) \in P$, we conduct one pass of backpropagation through the GNN to obtain gradients for the weights $\mathbf{W}^{l}$ using an appropriate loss function, e.g, (B)CE loss for (binary) classification. For each $\mathbf{W}^{l}$, the gradients obtained for each data batch are accumulated and %
represented as $\mathbf{G}^{l}$, which is defined as
\begin{equation}
    \begin{aligned}
            \mathbf{G}^{l} =   
\sum_{(\mathbf{t},\mathbf{X}) \in P}
        \frac{\partial\mathcal{L}\Big(  \Phi \big( \mathbf{X}; \; \{\mathbf{W}^{l}\}_{l=1}^{L} \big)\Big), \mathbf{t} \Big)}{\partial\mathbf{W}^{l}}.
    \end{aligned}
    \label{eq:make_gradients2}
    \nonumber
\end{equation}
For each layer, we select the set of honeypot neurons $N_{k}^{l}$ from all neurons $N$ by choosing the indices of the top-$k$ largest sums of accumulated absolute neuron gradients, where $k = n^{l} \cdot p$. Here, $n^{l}$ is the number of neurons in the layer, $p$ is the base percentage of honeypot neurons, and $axis=1$ indicates per-neuron summation. %
Note, that access to unlabeled data is a conservative assumption. Labeled data could enhance the defense by improving gradient-based identification of vulnerable weights. Notably, BFAs also rarely use labeled data; IBFA, the only GNN-specific BFA, avoids labeled data entirely~\cite{kummer2023ibfa}.

\begin{equation}
    \begin{aligned}
           N_{k}^{l} = \left\{ \argtopk_{} \left(\sum_{axis=1}|\mathbf{G}^{l}|\right) \right\} 
    \end{aligned}
    \label{eq:topk}
    \nonumber
\end{equation}

\paragraph{Choice of Scaling Parameters}
\citet{liu2023neuropots} highlight that neuron rescaling can introduce quantization errors, with each neuron's unique weight distribution requiring tailored rescaling to minimize these errors. Such quantization-induced perturbations %
propagate through matrix multiplications
, with their impact on network performance depending on each neuron's contribution to classification~\cite{zengperturbation,lenetperturbation}.
A uniform scaling parameter for all honeypots, as suggested by \citet{liu2023neuropots}, is thus suboptimal. 

We propose individualized scaling for each neuron to better minimize quantization error and preserve classification performance.
Initially, we augment the scaling factor $\gamma$ proportionally to the network's depth using a scaling factor $\lambda \geq 1$ to obtain %
a layer's scaling $\gamma^{l} = \gamma \cdot \lambda^{l}$.
This approach offers two benefits. First, it generally reduces propagated error as deeper layers are scaled higher. In GNNs, this favors distinguishing between nodes based on local neighborhoods, %
as it directs attackers towards deeper layers. Given that the evaluated GNNs may be shallower than the graph diameter, this strategy is consistent with using shallow architectures to prevent overfitting~\cite{ijcai2021p0618} and oversmoothing~\citep{liu2020towards}.

Further, we do not use $\gamma^{l}$ directly to rescale the honey neurons. Instead, we compute a saliency vector $\bar{\mathbf{s}}^{l}$ for each layer that considers the accumulated gradients magnitude
\begin{equation}
    \begin{aligned}
        \bar{\mathbf{s}}^{l} = 1+\frac{(\mathbf{s}^{l}-\min(\mathbf{s}^{l}))\cdot(\gamma^{l}-1)}{\max(\mathbf{s}^{l})-\min(\mathbf{s}^{l})} \text{, } \mathbf{s}^{l} = \sum_{axis=1}|\mathbf{G}^l_{N_{k}^{l}}|.%
    \end{aligned}
    \nonumber
\end{equation}
The saliency vector $\mathbf{s}^{l}$ is then multiplied with the honeypots $N_{k}^{l}$ using per-column division, $\mathbf{W}_{q, N_{k}^{l}}^{l} \oslash_1 \bar{\mathbf{s}}^{l}$ and the inputs are scaled using per-row product $\mathbf{X}_{N_{k}^{l}}^{l} \odot_0 \bar{\mathbf{s}}_ {l}$. %
\begin{figure*}%
    \centering %
    \includegraphics[width=1.0\textwidth, trim={0.1cm 0.25cm 17.5cm 0.25cm},clip]{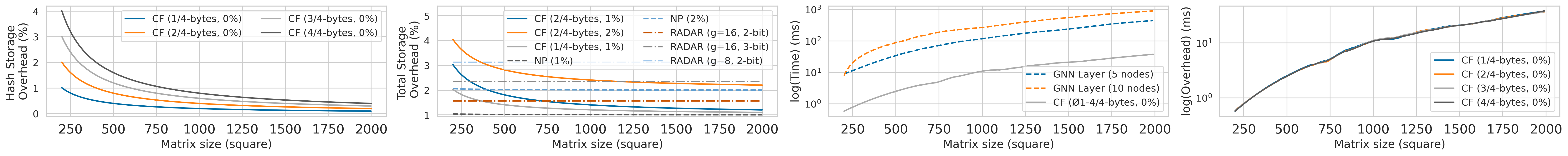}%
    \caption{\label{fig:crossfire_hash_all} 
    Storage overhead of Crossfire (CF) hashes relative to matrix size (\texttt{INT8}), varying cross digest sizes (left). Storage overhead of CF relative to matrix size, varying cross digest sizes and honeypot percentages (\%) compared to RADAR and NeuroPots (NP) (center%
    ). 
    Average hashing times (milliseconds) for CF across different cross digest sizes, plotted against the time complexity of a simple \texttt{INT8} GNN layer (%
    right) for 5 and 10 node graphs. %
    Layer digest sizes fixed at 4 bytes. %
    }%
\end{figure*}
\paragraph{Out of Distribution Weights}
Quantization can lead to limited range, i.e. for \texttt{INT8} (stored in two's complement) $\exists L^{l}, U^{l}, -128 \leq L^{l} \leq \min(\mathbf{W}_{q}^{l}) \wedge \max(\mathbf{W}_{q}^{l}) \leq U^{l} \leq 127$. BFAs typically target zero or near-zero weights and flip the most significant bit (MSB) of such weights~\cite{he2020defending2}, %
which has a high likelihood of creating an %
OOD element within that particular weight tensor. That is, if we choose $L^{l} = \min(\mathbf{W}_{q}^{l})$ and $U^{l} = \min(\mathbf{W}_{q}^{l})$, it holds that $\forall \widehat{w}_{ij} \in \widehat{\mathbf{W}}_{q}^{l}, w_{ij} \in \mathbf{W}_{q}^{l}| L^{l} > \widehat{w}_{ij} \vee U^{l} < \widehat{w}_{ij} \Rightarrow \widehat{w}_{ij} \neq w_{ij}$, indicating that $\widehat{w}_{ij}$ was subjected to a bit flip pushing it out of  $\mathbf{W}_{q}^{l}$ range. Under the assumption that double flips in a single $\widehat{w}_{ij}$ are uncommon and that an attacker flips the (or one of the) MSBs, potentially iteratively unsetting MSB-n bits in $\widehat{w}_{ij}$ until $L^{l} \leq \widehat{w}_{ij} \leq U^{l}$ will either entirely undo the attacker's bit flips or at least reduce their effect. For example, if we observe $L^{l} = -50$ (\texttt{11001110}), $U^{l} = 60$ (\texttt{00111100}) and $\widehat{w}_{ij} = -58$ (\texttt{11000110}), iteratively unsetting $\widehat{w}_{ij}$'s two MSB's will lead to $70$ (\texttt{01000110}) and $6$ (\texttt{00000110}), which, under the assumption an attacker preferably flips MSBs in near zero weights, could be a correct reconstruction for $w_{ij}$. Naturally, this approach has its own limitations, as, e.g. $\widehat{w}_{ij} = 14$ (\texttt{00001110}) would neither have been detected nor corrected for above $U^{l}, L^{l}$. However, as it is only complementary to our honeypots and checksum-based detection approach, it allows for a degree of reconstruction that would not be possible by simply replacing non-honeypot weights by zeros~\cite{li2021radar} or statistical approximations of $w_{ij}$~\cite{he2020defending2}.

\paragraph{Attack Detection and Reconstruction}
For attack detection and network verification, we opted for \texttt{Blake2b}~\cite{aumasson2013blake2}, a widely recognized hash function with established security analyses~\cite{guo2014analysis, rao2019comparative} and documented resource constraints~\cite{sugier2017memory}. %
\texttt{Blake2b} excels in speed and security, outperforming SHA-3. Its parallel compression and adaptable parameters make it versatile for various applications, confirming its importance in contemporary cryptographic practices.
Specifically, we compute the $d$-byte hashes for each row and column of a matrix $\mathbf{W}_{q}^{l}$ of size $n \times m$ with $d = 2$
\begin{align*}
\text{ColumnHash}(j, d) &= \texttt{Blake2b}\left( \sum_{i=1}^{m} \mathbf{W}_{q, ij}^{l} \right) \\
\text{RowHash}(i, d) &= \texttt{Blake2b}\left( \sum_{j=1}^{n} \mathbf{W}_{q, ij}^{l} \right)
\end{align*}
and store them in vectors $\mathbf{rh}^l, \mathbf{ch}^l$. The storage overhead for this is negligible relative to the storage costs of the network. If desired, $d$ (to which we, in this context, refer to as \emph{cross digest}) can also be chosen dynamically by, e.g., matrix size, $d = \min\left(\max\left(1, \frac{\log_2(n \cdot m)}{8}\right), M\right)$ with $M$ denoting the desired maximum number of bytes. %
Our hashing approach allows us to pinpoint changes to single elements in the matrix and surgically correct or remove them. This provides an advantage over RADAR, which can only detect changes in an entire group of weights and subsequently zeros them out, and it surpasses NeuroPots by enabling the detection of changes in non-honeypot weights. %
If we compute $\mathbf{rh}^l, \mathbf{ch}^l$ for $\mathbf{W}_{q}^{l}$ and, at some point after an attack on $\mathbf{W}_{q}^{l}$, obtain $\widehat{\mathbf{rh}}^l, \widehat{\mathbf{ch}}^l$ for $\widehat{\mathbf{W}}_{q}^{l}$, then $\widehat{\mathcal{R}}^l = \{ i \mid i \in [1, n], \mathbf{rh}^l_i \neq \widehat{\mathbf{rh}}^l_i \}$ and $\widehat{\mathcal{C}}^l = \{ i \mid i \in [1, m], \mathbf{ch}^l_i \neq \widehat{\mathbf{ch}}^l_i \}$ will be the row and column indices of perturbed elements in $\widehat{\mathbf{W}}_{q}^{l}$. For the elements located at these indices, 
repair mechanisms are executed. They are either replaced with honeypot weights (if they were a honeypot), have their distribution repaired by unsetting MSBs (if they were OOD), or are zeroed if none of the other options leads to positive %
layer verification. Note the initial sparsity induction via pruning increases the likelihood that zeroing resets weights to their pre-attack state.

\paragraph{Post Reconstruction Verification}
All weight matrices are hashed using the \texttt{Blake2b} hash function with a digest size of 4 bytes (referred to as \emph{layer digest}). This setup detects changes in the weights matrix with high reliability, as we demonstrate experimentally below. Although %
\texttt{Blake2b} with a 4-byte digest is more computationally expensive than simpler hashes %
used by, e.g., RADAR, it is only executed to detect an attack, triggering Crossfire's simpler, more granular mechanisms, and to verify post-attack reconstruction.

\begin{table}%
\centering
\caption{\label{tab:ogb_dataset} Overview of the six %
OGB \citep{hu2020open} benchmark \texttt{ogbg-mol} datasets %
used: %
number of graphs and tasks, average number of nodes, edges and recommended %
metric.}
\begin{tabular}{@{}llllll@{}}
\toprule
\bf Name & \bf Graphs & \bf Nodes & \bf Edges & \bf Tasks & \bf Metric\\ %
\midrule %
\texttt{pcba} & 437,929 & 26.0 & 28.1 & 128 & AP \\
\texttt{muv} & 93,087 & 24.2 & 26.3 & 17 & AP \\
\texttt{hiv} & 41,127 & 25.5 & 27.5 & 1 & AUROC \\
\texttt{toxcast} & 8,576 & 18.5 & 19.3 & 617 & AUROC \\
\texttt{tox21} & 7,831 & 18.6 & 19.3 & 12 & AUROC \\
\texttt{bace} & 1,513 & 34.1 & 36.9 & 1 & AUROC \\ %
\bottomrule
\end{tabular}
\end{table}
\section{Experiments} 
Our experiments\footnote{%
\url{https://github.com/lorenz0890/crossfireaaai2025}
} are designed to test the following hypotheses regarding Crossfire's performance on GNNs: \textbf{(a)} Crossfire has a higher chance of detecting bit flips compared to other methods, \textbf{(b)} Crossfire can repair the GNN so that it is indistinguishable from its pre-attack state, \textbf{(c)} Crossfire results in a smaller difference between pre-attack and post-reconstruction GNN performance, and \textbf{(d)} Crossfire does not %
increase the vulnerability of GNNs to evaluated BFAs.

We evaluate Crossfire against basic PBFA as well as IBFA, an attack specifically designed for GNNs with no known defense.
In the context of our experiments, an attack is considered detected if a single bit flip is identified by the detection mechanism of the respective defense strategy (i.e., RADAR, NeuroPots, or Crossfire). A network is considered reconstructed if applying the \texttt{Blake2b} hash function with a 4-byte digest to all weight matrices indicates they are identical to their pre-attack state. We optimize Crossfire's and NeuroPots' hyperparameters via grid search for the number of honeypots $p \in \{0.01, 0.05, 0.1\}$ (given as percentage of the number of neurons) and the (initial) rescaling factor $\gamma \in \{1.33, 1.66, 2.0\}$. We fix Crossfire's depth rescaling parameter at $\lambda = 1.1$ and \( \ell_1 \) pruning ratio at 75\% and compute gradients from 10 random samples from each dataset's training split. Note that, while the assumption of having unlabeled data falling into the same distribution as the original training data may not always hold in practice, selecting samples approximating the original training distribution might be a feasible substitute. 
For RADAR, we used group size $16$ and $2$-bit hashes, as recommended by its authors.

\subsubsection{Datasets and  Models} %
The same six Open Graph Benchmark (OGB) graph classification benchmark datasets are chosen %
for evaluation as in~\citet{kummer2023ibfa}'s work on IBFA. %
The goal in each dataset 
is to predict properties based on molecular graph structures, such that the datasets are consistent with the underlying 
assumptions of IBFA described by~\citet{kummer2023ibfa}. The %
datasets are split using a scaffold-based strategy, to ensure structural diversity between subsets~\citep{wu2018moleculenet, hu2020open}. 
Dataset characteristics are summarized in Table~\ref{tab:ogb_dataset}. Area under the receiver operating characteristic curve (AUROC) or average precision (AP) %
are used to measure prediction quality, %
as recommended by~\citet{hu2020open}.

To remain consistent with \citet{kummer2023ibfa}, %
we use 5-layer Graph Isomorphism Networks (GIN) quantized to \texttt{INT8} via scale quantization. GIN is widely adopted, integrated into popular frameworks \citep{Fey/Lenssen/2019}, and applied in research%
~\citep{
wang2023dynamic, 
gao2022malware}. See the appendix for details on datasets, GIN, and training.
\begin{figure*}[!h]%
    \centering
    \includegraphics[width=1.0\textwidth, trim={12.cm 0.25cm 0.1cm 0.25cm},clip]{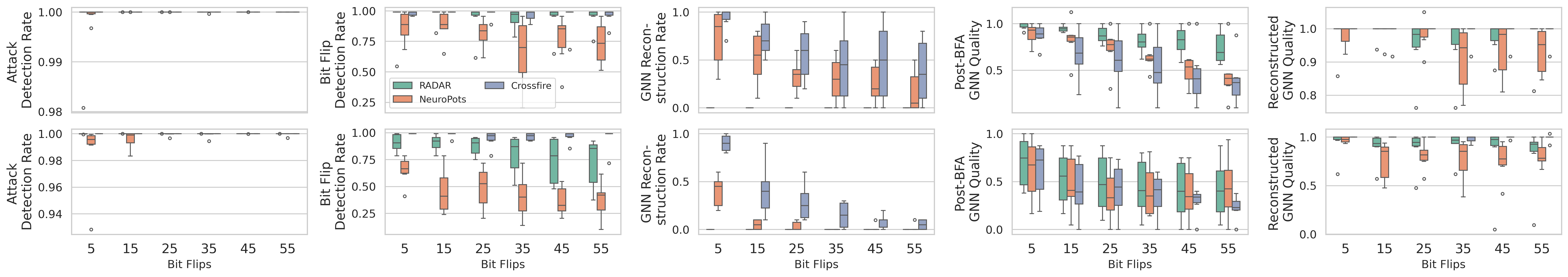}%
     \caption{\label{fig:detrates} Post-BFA and reconstructed GNN prediction qualities (with pre-attack normalized to 100\% to account for different scale quality metrics AP and AUROC), reconstruction, and detection rates for GNNs under PBFA (top row) and IBFA (bottom row) with varying amounts of bit flips, defended by NeuroPots, RADAR and Crossfire.  In each subplot, a box represents data aggregated from 6 datasets, with 10 runs per dataset, totaling 1080 runs for each IBFA and PBFA at optimal hyperparameters. We provide extensive tabular results in the technical appendix for completeness.%
     }
\end{figure*}
\subsection{Results}
We conducted an extensive evaluation of Crossfire, consisting of 2,160 individual experiments across six datasets, using a system with a NVIDIA %
GTX 1650 GPU (4GB VRAM)
and an Intel(R) %
i7-10750H CPU (64GB RAM).

\paragraph{Attack Detection and Mitigation}
For basic PBFA, Crossfire, NeuroPots, and RADAR achieved nearly 100\% attack detection across all datasets. Crossfire demonstrated superior detection of individual bit flips, outperforming competitors in %
most cases (Figure~\ref{fig:detrates}). Notably, Crossfire repaired damage to the GNN in over 50\% of the cases for up to 25 bit flips, significantly outperforming NeuroPots. By design, RADAR cannot reverse bit-flip-induced damage to restore a GNN to its pre-attack state. While selecting honeypots based on gradients could theoretically guide BFAs toward vulnerable areas and potentially lower post-attack GNN prediction quality, this effect was not significant (see below). Crossfire compensates with exceptional restoration performance, reliably restoring %
prediction quality to 100\% for up to 25 bit flips and outperforming the best competing method, which reached 98.6\%. Crossfire also restored prediction quality up to 98.1\% for up to 55 PBFA-induced bit flips.

Regarding IBFA (Figure~\ref{fig:detrates}), Crossfire's performance is slightly degraded compared to the PBFA attacker but still outperforms competitors by a large margin. %
While attack detection is near 100\% for all defenses, only Crossfire detects over 90\% of bit flips across datasets and restores over 99\% of the GNN's prediction quality for up to 55 bit flips. %
Moreover, Crossfire's reconstruction capabilities far exceed its weaker competitors. 
For example, for 15 bit flips, Crossfire achieves a 41\% average reconstruction rate, well outperforming its closest competitor, NeuroPots, which only achieves 5\%.
This trend of superiority is maintained for up to 55 bit flips.
Note that, while full recovery for large numbers of bit flips is modest (though significantly better than baselines), near-100\% attack detection probabilities for fewer flips makes a BFA unlikely to cause undetected damage, as inducing 25 consecutive flips would already require hours~\cite{wang2023aegis}.

To formalize our findings, we conducted Welch's t-test to compare Crossfire's performance with that of NeuroPots and RADAR, testing the hypotheses formulated at the beginning of this section for $p<10^{-3}$. For hypothesis \textbf{(a)}, we found a highly significant improvement in Crossfire's bit flip detection ratio ($t=9.8$, $p=10^{-18}$). For hypothesis \textbf{(b)}, Crossfire significantly outperformed its competitors in GNN reconstruction ratio ($t=7.1$, $p=2 \cdot 10^{-10}$). Regarding hypothesis \textbf{(c)}, Crossfire yielded a significantly smaller difference between pre-attack and post-reconstruction AP/AUROC ($t=4.7$, $p=5 \cdot 10^{-6}$). For hypothesis \textbf{(d)}, we found no significantly increased vulnerability in Crossfire-defended GNNs to the evaluated BFAs ($t=2.8$, $p=5 \cdot 10^{-3}$).

\paragraph{%
Overhead and Reliability}
For estimating relative computational overhead, we assume an \texttt{INT8} quantized simplistic message passing network operating over adjacency matrix $\mathbf{A}$, feature matrix $\mathbf{X}$ and weight matrix $\mathbf{W}$, computing $\mathbf{A}\mathbf{X}\mathbf{W}^{T}$. We evaluate for batch size 32 and a randomly connected graph with 5 or 10 nodes (i.e. $\mathbf{A} \in \{0,1\}^{(5 \times 5)}$ or $\mathbf{A} \in \{0,1\}^{(10 \times 10)}$). In practice, the number of nodes is higher (Table~\ref{tab:ogb_dataset}), so overhead can be assumed to be even lower. We run hashing sequentially to simulate the worst case of a system where parallelism is not possible. %
As shown in Figure~\ref{fig:crossfire_hash_all}, right, Crossfire
requires one order of magnitude less computation time than even our simplistic GNN layer. The average runtime stays low 
even for large matrices %
and decreases relative to the GNN layer's computational demand as matrix sizes increase, demonstrating Crossfire's scalability.
Comparing the CPU-based hashing with the \texttt{INT8} quantized GNN's complexity is appropriate, as \texttt{INT8} quantized models are typically run on CPUs on edge devices. The storage overhead scales effectively with increasing matrix sizes (Figure~\ref{fig:crossfire_hash_all}, left). Depending on the %
digest sizes and honeypot percentages, Crossfire was more efficient than NeuroPots and RADAR (Figure~\ref{fig:crossfire_hash_all}, center).

To test \texttt{Blake2b}'s reliability for post-attack verification, we induced 1, 5, and 10 random consecutive bit flips in uniformly initialized square \texttt{INT8} matrices of sizes ranging from 100 to 1000 (in increments of 100). We then tested the probability that a bit flip goes undetected for layer digest sizes of 1, 2, and 3. The experiment was repeated 100 times for each combination of the above parameters. We found that a digest size of 1 failed to detect 0.66\%, 0.44\%, and 0.22\% of the 1, 5, and 10 bit flip sequences, respectively. However, any digest size $\geq 2$ achieved a 100\% detection rate, highlighting \texttt{Blake2b}'s usefulness for integrity verification.

\section{Conclusion}
We introduced Crossfire, the first defense framework designed to protect GNNs from BFAs which is robust and retraining-free. Tested on six datasets accross 2,160 experiments, Crossfire surpasses existing methods in both detection and recovery performance. Crossfire achieves near-perfect attack and bit-flip detection and typically restores prediction quality to pre-attack levels. %
On average, Crossfire offers a 21.8\% higher probability of full reconstruction %
and improves post-repair prediction quality by 10.85\% over its competitors. Crossfire's computational and storage overhead is negligible compared to the inherent complexity of the simplest GNNs. The innovative use of hashing and honeypots, combined with leveraging 
sparsity and the systematic unflipping of bits in OOD weights, enables identification and repair of perturbed elements with minimal overhead. Crossfire addresses a critical gap in GNN security, offering a scalable solution to safeguard GNNs in adversarial scenarios.

\section{Acknowledgments}
This work was supported by the Vienna Science and Technology
Fund (WWTF) [10.47379/VRG19009].%

\bibliography{aaai25}

\appendix

\begin{figure*}
    \centering
    \includegraphics[width=1.0\textwidth, trim={0.1cm 0.25cm 0.1cm 0.25cm},clip]{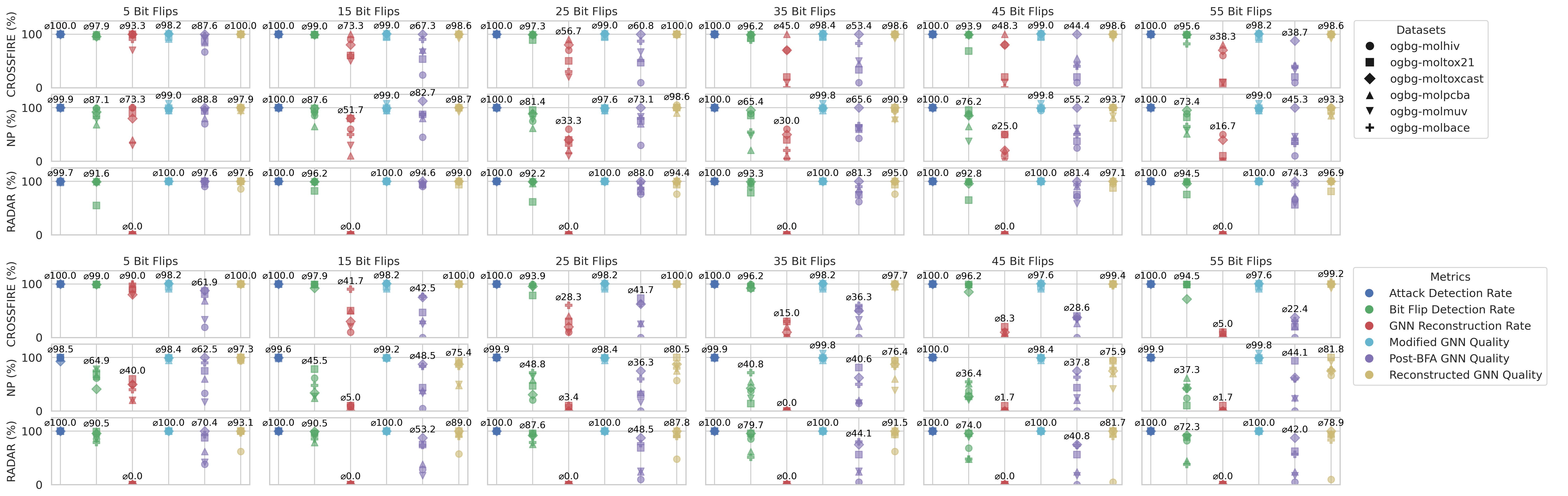}%
     \caption{\label{fig:detrates_detail} Pre- and post-BFA GNN prediction qualities (with pre-attack normalized to 100\% to account for different scale quality metrics AP and AUROC), reconstruction, and detection rates for GNNs under PBFA (top three rows) and IBFA (bottom three rows) with varying amounts of bit flips, defended by NeuroPots (NP), RADAR, and Crossfire. 
     Each data point represents per-dataset averages of 10 runs per dataset, totaling 1080 runs for each IBFA and PBFA at optimal hyperparameters, the average of each column is displayed on top of its maximal value. }
\end{figure*}

\begin{table*}
    \centering
    \caption{Tabular results for IBFA on a 5-layer GIN defended by NeuroPots (NP), RADAR, and Crossfire using optimized hyperparameters across six \texttt{ogbg-mol} datasets with 5 to 55 bit flips. The table presents the AP/AUROC of the original model, post-BFA, and post-reconstruction, along with the reconstruction ratio (Rec.), bit flip detection ratio (Flip Det.), and attack detection ratio (BFA Det.). The results are aggregated means over all experiments (5 to 55 bit flips; see Figure\ref{fig:detrates_detail}). The best results per row are marked in \textbf{bold} for each dataset.}

    \label{tab:ibfa_results_table}
    \begin{tabular}{llccccccc}
\toprule
        \textbf{Dataset} & \textbf{Algorithm} & \textbf{AP/AUROC } & \textbf{AP/AUROC (BFA)} & \textbf{AP/AUROC (Rec.)} & \textbf{Flip Det.} & \textbf{Rec.} & \textbf{BFA Det.} \\
\midrule
\texttt{bace} & Crossfire &     0.80 &           0.67 &            \textbf{0.80} &      \textbf{0.99} & \textbf{0.48} &     1.00 \\
\texttt{bace} &        NeuroPots &     0.80 &           0.70 &            0.75 &      0.60 & 0.10 &     1.00 \\
\texttt{bace} &     RADAR &     0.80 &           \textbf{0.73} &            0.77 &      0.63 & 0.00 &     1.00 \\
\midrule
\texttt{hiv} & Crossfire &     0.71 &           0.50 &            \textbf{0.71} &      \textbf{0.99} & \textbf{0.18} &     1.00 \\
\texttt{hiv} &        NeuroPots &     0.71 &           0.51 &            0.68 &      0.40 & 0.10 &     1.00 \\
\texttt{hiv} &     RADAR &     0.71 &           \textbf{0.53} &            0.58 &      0.85 & 0.00 &     1.00 \\
\midrule
\texttt{muv} & Crossfire &     0.12 &           \textbf{0.03} &            \textbf{0.12} &      \textbf{0.97} & \textbf{0.20} &     1.00 \\
\texttt{muv} &        NeuroPots &     0.12 &           \textbf{0.03} &            0.08 &      0.43 & 0.03 &     1.00 \\
\texttt{muv} &     RADAR &     0.12 &           \textbf{0.03} &            \textbf{0.12} &      0.88 & 0.00 &     1.00 \\
\midrule
\texttt{pcba} & Crossfire &     0.21 &           0.06 &            0.19 &      \textbf{0.99} & \textbf{0.38} &     1.00 \\
\texttt{pcba} &        NeuroPots &     0.21 &           \textbf{0.07} &            0.14 &      0.52 & 0.03 &     1.00 \\
\texttt{pcba} &     RADAR &     0.21 &           \textbf{0.07} &            \textbf{0.20} &      0.68 & 0.00 &     1.00 \\
\midrule
\texttt{tox21} & Crossfire &     0.66 &           0.58 &            0.65 &      0.94 & \textbf{0.38} &     1.00 \\
\texttt{tox21} &        NeuroPots &     0.66 &           0.60 &            \textbf{0.66} &      0.41 & 0.17 &     1.00 \\
\texttt{tox21} &     RADAR &     0.66 &           \textbf{0.61} &            \textbf{0.66} &      \textbf{0.95} & 0.00 &     1.00 \\
\midrule
\texttt{toxcast} & Crossfire &     0.58 &           0.55 &            \textbf{0.58} &      0.89 & \textbf{0.25} &     1.00 \\
\texttt{toxcast} &        NeuroPots &     0.58 &           0.56 &            0.57 &      0.37 & 0.08 &     0.99 \\
\texttt{toxcast} &     RADAR &     0.58 &           \textbf{0.57} &            \textbf{0.58} &      \textbf{0.95} & 0.00 &     1.00 \\
\bottomrule
\end{tabular}
\end{table*}

\begin{table*}
    \centering
    \caption{Tabular results for PBFA on a 5-layer GIN defended by NeuroPots (NP), RADAR, and Crossfire using optimized hyperparameters across six \texttt{ogbg-mol} datasets with 5 to 55 bit flips. The table presents the AP/AUROC of the original model, post-BFA, and post-reconstruction, along with the reconstruction ratio (Rec.), bit flip detection ratio (Flip Det.), and attack detection ratio (BFA Det.). The results are aggregated means over all experiments (5 to 55 bit flips; see Figure\ref{fig:detrates}). The best results per row are marked in \textbf{bold} for each dataset.}
    \label{tab:pbfa_results_table}
    \begin{tabular}{llccccccc}
\toprule
        \textbf{Dataset} & \textbf{Algorithm} & \textbf{AP/AUROC}  & \textbf{AP/AUROC (BFA)} & \textbf{AP/AUROC (Rec.)} & \textbf{Flip Det.} & \textbf{Rec.} & \textbf{BFA Det.} \\
\midrule
\texttt{bace} & Crossfire &     0.80 &           0.72 &            \textbf{0.80} &      0.94 & 0.30 &     1.00 \\
\texttt{bace} &        NeuroPots &     0.80 &           0.71 &            0.79 &      0.81 & \textbf{0.35} &     1.00 \\
\texttt{bace} &     RADAR &     0.80 &           \textbf{0.78} &            \textbf{0.80} &     \textbf{ 0.97} & 0.00 &     1.00 \\
\midrule
\texttt{hiv} & Crossfire &     0.71 &           0.55 &            \textbf{0.71} &      \textbf{0.99} & \textbf{0.78} &     1.00 \\
\texttt{hiv} &        NeuroPots &     0.71 &           0.58 &            0.70 &      0.88 & 0.63 &     1.00 \\
\texttt{hiv} &     RADAR &     0.71 &           \textbf{0.66} &            0.69 &      0.98 & 0.00 &     1.00 \\
\midrule
\texttt{muv} & Crossfire &     0.12 &           0.07 &            0.11 &      0.98 & \textbf{0.28} &     1.00 \\
\texttt{muv} &        NeuroPots &     0.12 &           \textbf{0.09} &            0.11 &      0.64 & 0.12 &     1.00 \\
\texttt{muv} &     RADAR &     0.12 &           \textbf{0.09} &            \textbf{0.12 }&      \textbf{0.99} & 0.00 &     1.00 \\
\midrule
\texttt{pcba} & Crossfire &     0.21 &           0.12 &            0.20 &      \textbf{0.99} & \textbf{0.95} &     1.00 \\
\texttt{pcba} &        NeuroPots &     0.21 &           0.13 &            0.18 &      0.55 & 0.17 &     1.00 \\
\texttt{pcba} &     RADAR &     0.21 &           \textbf{0.19} &            \textbf{0.21} &      0.98 & 0.00 &     1.00 \\
\midrule
\texttt{tox21} & Crossfire &     0.66 &           0.57 &            0.65 &      \textbf{0.90} & 0.43 &     1.00 \\
\texttt{tox21} &        NeuroPots &     0.66 &           0.60 &            \textbf{0.66} &      \textbf{0.90} & \textbf{0.52} &     1.00 \\
\texttt{tox21} &     RADAR &     0.66 &           \textbf{0.63} &            0.65 &      0.69 & 0.00 &     1.00 \\
\midrule
\texttt{toxcast} & Crossfire &     0.58 &           \textbf{0.58} &            \textbf{0.58} &      \textbf{0.98 }& \textbf{0.80} &     1.00 \\
\texttt{toxcast} &        NeuroPots &     0.58 &           \textbf{0.58} &            \textbf{0.58} &      0.93 & 0.52 &     1.00 \\
\texttt{toxcast} &     RADAR &     0.58 &           \textbf{0.58} &            \textbf{0.58} &      \textbf{0.98} & 0.00 &     1.00 \\
\bottomrule
\end{tabular}
\end{table*}

\section{Appendix A: Graph Isomorphism Networks}

The Graph Isomorphism Network (GIN) possesses the same ability as the Weisfeiler-Lehman (1-WL) test to distinguish between non-isomorphic graphs~\citep{leskovec2019powerful}. While extensive research aims to enhance the expressivity of GNNs) beyond this level~\citep{wlsurvey}, neighborhood aggregation remains a prevalent technique in practical applications. The 1-WL test is generally adequate for differentiating most graphs in widely used benchmark datasets~\citep{Zopf22a, morris2021power}. According to~\citet{leskovec2019powerful}, a neighborhood aggregation GNN with a sufficient number of layers can achieve the discriminative power of the 1-WL test if the aggregation ($\Agg$) and combination ($\Comb$) functions in each layer's update rule, as well as its graph-level readout ($\Readout$), are injective.
GIN achieves this with the update rule
\begin{equation}
    \mathbf{h}_v^{k} = \MLP^{k}\left ( (1+\epsilon^{k} ) \cdot \mathbf{h}_v^{k-1} + \sum_{u \in N(v) } \mathbf{h}_u^{k-1} \right )
    \label{eq:defgin}
\end{equation}
integrating a multi layer perceptron (MLP) into  $\Comb^{k}$, realizing a universal function approximator on multisets~\citep{ZaheerKRPSS17, hornik1991approximation, hornik1989multilayer}.
When input features are represented as one-hot encodings, an MLP is unnecessary before summation in the first layer, as the summation itself is injective in this scenario. In GIN, the graph-level readout is accomplished by first summing the embeddings from each layer and then concatenating (denoted by $\Vert$) these summed vectors across the layers of the GIN. The resulting graph-level embedding $\mathbf{h}_G$ can subsequently be used as input for tasks such as graph-level classification, potentially through an MLP.
\begin{equation}
    \mathbf{h}_G = \Big\Vert_{k=0}^n \left ( \sum_{ v \in V(G) } \mathbf{h}_v^{k} \right )
    \label{eq:defginreadout}
\end{equation} 

\section{Appendix B: Quantized Models}
We trained \texttt{INT8} quantized models on each dataset's training split using the Straight-Through Estimator (STE) method, as described in our main paper. 
The Adam optimizer was useed with a learning rate of $10^{-3}$, and models were trained for 30 epochs. While more advanced models and quantization methods could potentially enhance prediction quality, our objective was not to surpass the current state-of-the-art in prediction accuracy, but to illustrate GIN's susceptibility to IBFA. It's important to note that some datasets involve highly challenging learning tasks, and our quantized GIN training results are comparable to those reported by OGB~\citep{hu2020open} for \texttt{FLOAT32} training

\section{Appendix C: Dataset Descriptions}
The task associated with the \texttt{ogb-molhiv} dataset is to predict whether a certain molecule structure inhibits human immunodeficiency virus (HIV) or not. In the larger \texttt{ogb-molpcba} dataset each graph represents a molecule, where nodes are atoms, and edges are chemical bonds, and the task is to predict 128 different biological activities (inactive/active). The \texttt{ogb-moltox21} dataset contains data with qualitative toxicity measurements on 12 biological targets. The \texttt{ogbg-toxcast} dataset is another toxicity related dataset. The \texttt{obgbg-molbace} dataset is a biochemical single task binary classification (inhibition of human $\beta$-secretase 1 (BACE-1)) dataset. The \texttt{ogbg-molmuv} dataset is a subset of PubChem BioAssay commonly used for evaluation of virtual screening techniques.

\section{Appendix D: Tabular Results}
The following section presents tabular results complementing the main paper's visualizations, detailing the performance of Crossfire, NeuroPots, and RADAR in detecting and reconstructing PBFA and IBFA attacks across six datasets.

Tables~\ref{tab:ibfa_results_table} and~\ref{tab:pbfa_results_table} summarize the key performance metrics, including attack detection rates, GNN reconstruction rates as well as AP/AUROC restoration levels, and an even more fine-granular view can be found in Figure~\ref{fig:detrates_detail}. Recall that, in our experiments, an attack is deemed detected if any individual bit flip is identified. A bit flip is considered detected when it is recognized by the detection system of the respective defense mechanism (i.e., RADAR, NeuroPots, or Crossfire). Finally, a network is classified as reconstructed if the Blake2b hash function with a 4-byte digest shows that all weight matrices match their pre-attack state.

For both IBFA and PBFA, all defense algorithms provide nearly perfect attack detection ratios. Crossfire outperforms its competitors in terms of reconstruction ratio after IBFA, as it consistently offers the best chance of restoring a GNN to its pre-attack state across all datasets. Regarding PBFA, Crossfire maintains strong performance, leading reconstruction rates in 66\% of the evaluated cases. Similarly, Crossfire demonstrates a favorable detection ratio for individual bit flips, outperforming or matching its competitors in 66\% of the cases as well. Compared to NeuroPots and RADAR, Crossfire achieves better-reconstructed AP/AUROC after PBFA or IBFA in 58\% of the evaluated cases, with its performance advantages being particularly pronounced when defending against IBFA.

To further formalize these findings, we performed Welch's t-test to compare Crossfire's bit flip detection ratio and GNN reconstruction ratios against NeuroPots and RADAR. The results indicate that for the bit flip detection ratio (t-statistic: $9.86$, p-value: $1.66\cdot10^{-18}$) and reconstruction ratio (t-statistic: $7.11$, p-value: $1.75\cdot10^{-10}$), Crossfire demonstrated a highly significant improvement over its competitors ($p < 1\cdot10^{-3}$).

Figure~\ref{fig:detrates_detail} offers a finer-grained visualization of the results, including the pre-attack prediction qualities of GNNs modified by NeuroPots and Crossfire (RADAR is designed not to modify the model it protects). Notably, Crossfire's modifications introduce quantization errors comparable to those of NeuroPots.

To summarize, we believe these results demonstrate that we have achieved our goal of developing a method capable of verified reconstruction of GNNs after BFA.
\end{document}